\documentclass[10pt,letterpaper]{article}

\usepackage{cogsci}
\usepackage{graphicx}
\usepackage{float}

 \cogscifinalcopy 

\usepackage[
  style=apa,
  natbib=true,
  annotation=false,
]{biblatex}
\usepackage{float} 

\title{Effort as Ceiling, Not Dial: Reasoning Budget Does
Not Modulate Cognitive Cost Alignment Between Humans and
Large Reasoning Models}

\author[1]{\mbox{Yueqing Hu (scnu.psy.hyq@gmail.com)}}
\author[2]{\mbox{Tianhong Wang}}
\affil[1]{Institute of Neuroscience, Chinese Academy of Sciences, Shanghai, China}
\affil[2]{School of Philosophy, Anhui University, Hefei, China}

\begin{document}

\maketitle

\begin{abstract}
Large Reasoning Models (LRMs) generate chain-of-thought traces
whose length tracks human reaction times across cognitive tasks,
but recent debate questions whether this alignment reflects
genuine computational structure or surface verbosity. We test
whether the alignment varies with inference-time reasoning
effort. Across GPT-OSS-20B and GPT-OSS-120B, three effort
levels, and six reasoning tasks, within-task and cross-task
alignment remain invariant: Bayes Factors lean toward the null,
and mean alignment is numerically near-identical across
conditions. A manipulation check reveals that the effort
parameter sets an upper budget on generation rather than driving
real-time allocation, suggesting that the allocation policy is
crystallized at training time. Arithmetic complexity contrasts
further show that token allocation tracks fine-grained,
format-dependent human difficulty patterns, with model scale
improving the match. Cognitive cost alignment between LRMs and
humans appears to be a training-time achievement, robust to
inference-time perturbations, supporting a compiled rather than
online account of LRM problem-solving.

\textbf{Keywords:}
large reasoning models;
chain-of-thought;
reasoning effort;
cognitive cost alignment;
effort invariance;
arithmetic cognition
\end{abstract}

\section{Introduction}

A central goal of cognitive science is to determine whether
artificial systems not only solve problems correctly, but do so
through processes analogous to human cognition
\citep{lake2017building, binz2023using}. The emergence of Large
Reasoning Models (LRMs), neural networks trained via Reinforcement
Learning with Verifiable Rewards (RLVR)
\citep{guo2025deepseek, lambert2024tulu} to generate step-by-step
chain-of-thought (CoT) before producing an answer
\citep{wei2022chain}, has opened a new avenue for testing this
question. \citet{de2025cost} reported a striking convergence:
across seven reasoning tasks varying in cognitive demand, the
length of CoT reasoning traces generated by DeepSeek-R1 reliably
predicted human response times (RTs), both within tasks (mean
$\bar{r} = 0.57$) and across tasks ($r = 0.97$). This pattern held
without any built-in symbolic mechanisms, suggesting that
goal-directed optimization may implicitly recover core features of
human problem-solving complexity
\citep{gershman2015computational}.

This finding has prompted three lines of skeptical inquiry.
\citet{vankov2026correlations} raised a causal concern, manipulating
reasoning effort in GPT-OSS-120B and finding negligible effects on
accuracy in five of six tasks, alongside lengthy traces for
trivially simple problems (e.g., ``$2 + 2 = \text{?}$''); they
concluded that token output may reflect learned verbosity templates
rather than dynamic resource allocation.
\citet{hu2026thinking} raised a faithfulness concern: token-level
traces may function as performative scaffolding that conditions
subsequent generations without faithfully representing the
underlying computations \citep{stechly2025beyond,
palod2025performative}. \citet{dujmovic2026no} raised a mechanistic
concern, arguing via a ``bird vs.\ car'' analogy that correlated
behavioral profiles between two systems do not license inferences
about shared algorithmic mechanisms. In a series of replies,
\citet{de2026vankov, de2026hu, de2026dujmovic} defended the
alignment as a robust empirical phenomenon at the rational rather
than algorithmic level, while acknowledging that further evidence
is needed.

These exchanges share a critical limitation: all six contributions
treat alignment as a binary property (present or absent), rather
than asking whether it varies \emph{as a function of} reasoning
effort. If alignment genuinely reflects computational structure, it
should remain stable across effort conditions, neither collapsing
under minimal effort nor inflating under maximal effort. Conversely,
if the observed correlations merely capture surface-level
verbosity, they may be selectively disrupted or enhanced by effort
manipulations in ways unrelated to human cognition
\citep{mccoy2024embers}.

Parallel evidence from reasoning distillation reinforces the
importance of this question. \citet{hu2026h} demonstrated that
Supervised Fine-Tuning (SFT) distillation, the dominant paradigm
for deploying efficient LRMs, induces a ``Functional Alignment
Collapse'': although distilled student models retain the
\emph{form} of extended reasoning traces, their alignment with
human cognitive costs degrades markedly relative to their
RLVR-trained teachers (from $\bar{r} = 0.64$ to $\bar{r} = 0.34$).
This indicates that human-like cost scaling is an emergent property
of active reinforcement learning, sensitive to training-time
perturbations. Whether it is equally sensitive to inference-time
perturbations, such as changes in the reasoning effort budget,
remains unknown.

The present study addresses this question directly. We evaluated
GPT-OSS-20B and GPT-OSS-120B \citep{agarwal2025gpt} under three
reasoning effort conditions (low, medium, high) on six reasoning
tasks adapted from \citet{de2025cost}, spanning arithmetic, formal
logic, relational reasoning, and intuitive judgment. We formalized
two substantive predictions and one null. The \textbf{Linear
Enhancement Hypothesis} (H1) predicts a monotonic increase in
alignment as effort increases, on the reasoning that greater
inference-time computation should more faithfully mirror the graded
difficulty structure of human cognition. The \textbf{Optimal Effort
Hypothesis} (H2) predicts an inverted-U pattern, with medium effort
yielding maximal alignment because low effort produces
underthinking and high effort produces overthinking relative to
human cognitive norms \citep{shenhav2013expected}. The
\textbf{Effort Invariance null} (H3) predicts that alignment will
remain stable across effort conditions, which would suggest that
the model's allocation policy is fixed at training time rather than
dynamically reconfigured at inference time
\citep{anderson1982acquisition, logan1988toward}.

To further address Vankov et al.'s challenge regarding arithmetic
specifically, we conducted a systematic complexity-contrast
analysis across four manipulations of arithmetic difficulty
(operation type, carry structure, operand count, and digit count),
comparing model token profiles to human RT profiles under each
effort setting. Our results converge on H3: reasoning effort has no
systematic effect on LRM and human alignment, and the alignment
that does obtain is structurally meaningful, tracking fine-grained
complexity manipulations in directions consistent with human
behavior.

\section{Methods}

\subsection{Datasets}

\begin{figure}[!tbp]
  \centering
  \includegraphics[width=\columnwidth]{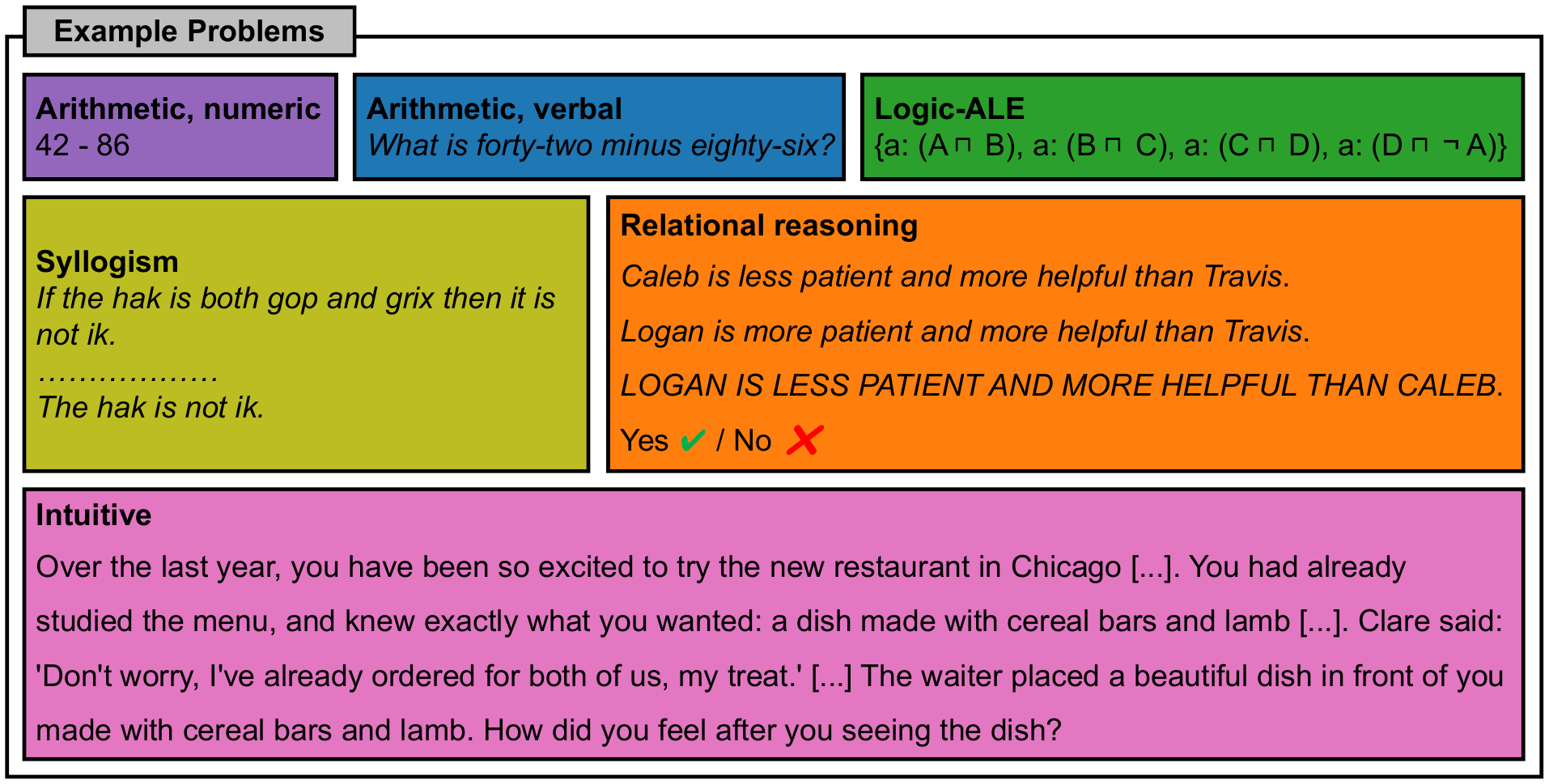}
  \caption{Example reasoning problems across the six tasks.}
  \label{fig:tasks}
\end{figure}

We utilized six reasoning tasks adapted from \citet{de2025cost},
spanning diverse cognitive domains (see Figure~\ref{fig:tasks}).
The H-ARC task from the original battery was excluded, as recent
work argues that ARC tasks depend fundamentally on visuospatial
priors rather than linguistic reasoning \citep{hu2025arc}. Full
task descriptions and human behavioral data are reported in
\citet{hu2026h}; we provide a brief overview here.

\begin{enumerate}
  \item \textbf{Arithmetic (numeric \& verbal):} 168 items (84
    per format) requiring rigid algorithmic computation, solved
    by $N=60$ participants.
  \item \textbf{Syllogisms:} 32 deductive logic problems, $N=24$.
  \item \textbf{Logic-ALE:} 20 consistency-judgment problems in
    Argument Logic Evaluation format, $N=84$.
  \item \textbf{Relational Reasoning:} 84 transitive inference
    items, $N=310$.
  \item \textbf{Intuitive Reasoning:} 144 causal-judgment
    vignettes, $N=58$.
\end{enumerate}

\subsection{Models and Reasoning Effort}

\begin{figure}[!tbp]
  \centering
  \includegraphics[width=\columnwidth]{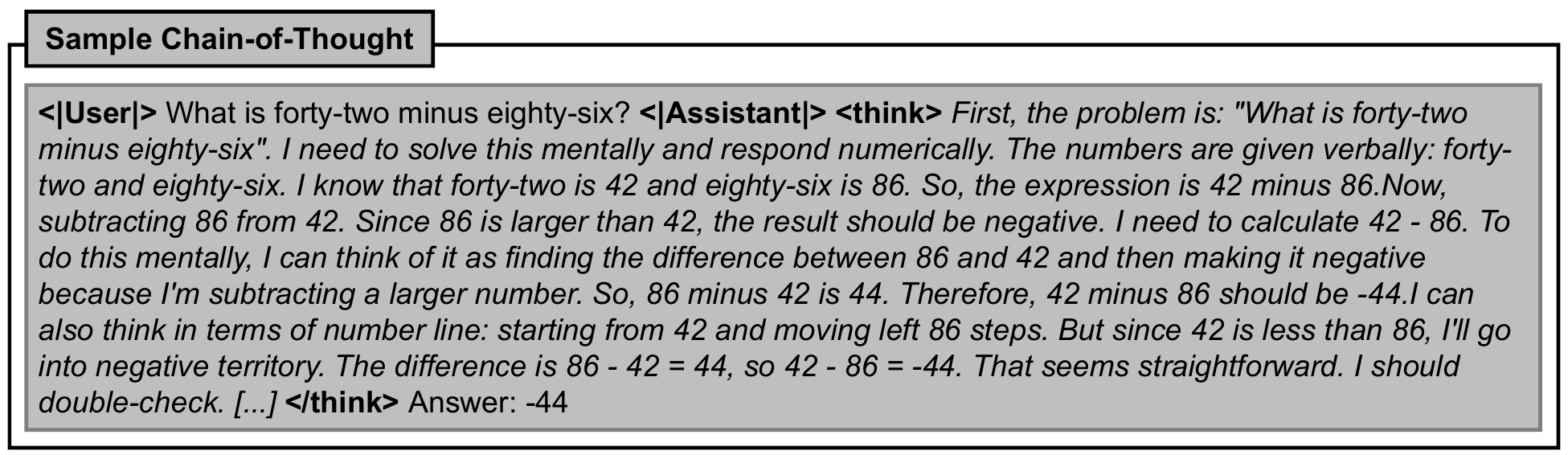}
  \caption{Sample chain-of-thought produced by GPT-OSS-120B
    (medium effort). The \texttt{<think>} block contains the
    reasoning trace whose token length constitutes our measure
    of reasoning cost.}
  \label{fig:prompt}
\end{figure}

We evaluated two open-weight LRMs from the GPT-OSS family:
\textbf{GPT-OSS-20B} and \textbf{GPT-OSS-120B}
\citep{agarwal2025gpt}. Both models support an explicit
\texttt{reasoning\_effort} parameter that modulates the length
budget allocated to the internal chain-of-thought, with three
discrete levels: \texttt{low}, \texttt{medium}, and \texttt{high}.
All other generation parameters were held fixed (temperature
$T = 0$, greedy decoding) to ensure deterministic
reproducibility. A representative output is shown in
Figure~\ref{fig:prompt}.

\subsection{Measures}

For each problem $i$, model $m$, and effort condition $e$, we
extracted:

\begin{itemize}
  \item \textbf{Accuracy ($Acc_{m,e,i}$):} Binary correctness,
    validated against ground truth.
  \item \textbf{Reasoning Cost ($C_{m,e,i}$):} Token count within
    the \texttt{<think>} delimiters, serving as the proxy for
    inference-time computational cost. Token count is a coarse
    proxy; more granular measures such as compute per token
    \citep{chen2026think} may better capture intrinsic reasoning
    effort, but token count remains the standard measure in this
    literature and enables direct comparison with
    \citet{de2025cost}.
\end{itemize}

\subsection{Analysis}

\paragraph{Manipulation check.}
Before interpreting any alignment statistic, we verified that
the \texttt{reasoning\_effort} parameter substantially modulated
the length of generated traces. For each (model $\times$ task)
cell, we quantified effort-induced variation in token counts as
the high-to-low fold change in mean token count,
$\Delta_{m,t} = \bar{C}_{m,\text{high},t} / \bar{C}_{m,\text{low},t}$,
with values near 1 indicating that the manipulation did not
reliably alter trace length. This step is essential: if effort
fails to modulate trace length, then an effort-invariant alignment
result is uninformative, as it would reflect a failed manipulation
rather than a genuine null. We therefore report manipulation-check
statistics alongside the substantive analyses, and stratify the
interpretation of invariance by whether the manipulation took
hold for a given task.

\paragraph{Within-task alignment.}
For each combination of model $m$, effort level $e$, and task
$t$, we computed the Pearson correlation between log-transformed
reasoning cost and log-transformed human RT:
\begin{equation}
  r_{m,e,t} = \text{Corr}\!\left(\log C_{m,e,t},\, \log RT_t\right)
\end{equation}
Log transforms were applied because both token counts and RTs are
positive, right-skewed quantities for which multiplicative
relationships are more natural than additive ones.

\paragraph{Cross-task alignment.}
We aggregated task-level mean token counts and human RTs and
computed a single cross-task Pearson correlation per
(model $\times$ effort) condition to capture how well token usage
mirrors broad differences in cognitive demand across task types.

\paragraph{Bayesian inference on effort invariance.}
To directly test whether effort modulates within-task alignment,
we conducted Bayesian paired-samples $t$-tests on Fisher
$z$-transformed $r$ values ($z = \text{arctanh}(r)$) across all
pairwise effort contrasts (low vs.\ medium, medium vs.\ high,
low vs.\ high). Sample size for each paired test was $n = 12$
(2 models $\times$ 6 tasks), which we acknowledge is modest;
Bayes Factors are therefore interpreted as one piece of
converging evidence rather than definitive. Bayes Factors
($\text{BF}_{10}$) were computed under the Jeffreys-Zellner-Siow
(JZS) prior with scale $r = 0.707$ \citep{rouder2009bayesian},
and interpreted following the conventional thresholds in
\citet{wagenmakers2018bayesian}: $\text{BF}_{10} < 1/3$ as moderate
evidence for the null, $1/3 \leq \text{BF}_{10} \leq 3$ as
anecdotal or inconclusive, and $\text{BF}_{10} > 3$ as moderate
evidence for the alternative.

\paragraph{Arithmetic complexity contrasts.}
Following \citet{de2026vankov}, we tested whether models exhibit
the same pattern of difficulty scaling as humans along four
within-task arithmetic manipulations: operation type (addition
vs.\ subtraction), carry structure (no-carry vs.\ carry), operand
count (2 vs.\ 3), and digit count (1 vs.\ 2). These four
dimensions reflect well-established sources of arithmetic
difficulty in the human cognitive literature
\citep{ashcraft1992cognitive, campbell2001cognitive}. For each
contrast, numeric and verbal formats were analyzed separately,
yielding 16 directional predictions per effort level. We
conducted Welch's $t$-tests comparing token counts between the
two conditions of each contrast, pooling across model sizes and
treating direction consistency (all $t < 0$, i.e., the more
complex condition generates more tokens) as the primary index
of structural alignment.

\section{Results}

\subsection{Manipulation Check: Effort Modulates Trace Length
Only Modestly}

We first verified whether the \texttt{reasoning\_effort}
parameter substantially modulated the length of generated
traces. The high-to-low fold change in mean token count,
$\Delta_{m,t}$, was small in most (model $\times$ task) cells.
For both models on Arithmetic (numeric and verbal), mean token
counts varied negligibly across effort levels
($\Delta \approx 1.0$; see Figure~\ref{fig:barchart}). The
manipulation produced larger absolute differences on tasks
with longer baseline traces (Logic-ALE, Relational), but even
there the variation across effort levels was small relative
to the order-of-magnitude differences across tasks. This
converges with the observation of
\citet{vankov2026correlations} that the
\texttt{reasoning\_effort} parameter has only limited
behavioral consequences. We interpret this as evidence that
the parameter functions less as a real-time computational
dial and more as an upper budget on how many tokens the model
\emph{may} generate, while the actual allocation across items
is largely determined by training-time policy. This
observation is itself informative, and we return to its
theoretical implications in the Discussion. For the analyses
below, we accordingly do not expect large effort-induced
shifts in alignment, but ask whether any residual modulation
of the parameter produces systematic shifts.

\subsection{Effort Does Not Disrupt Within-Task Alignment}

\begin{figure*}[!tbp]
  \centering
  \includegraphics[width=\textwidth]{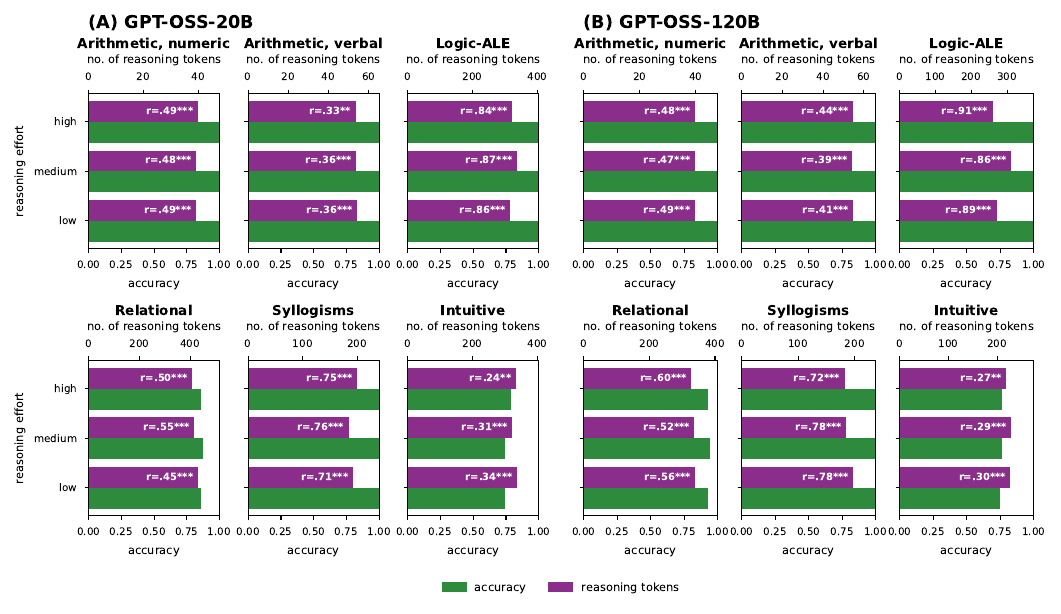}
  \caption{\textbf{Within-task alignment across reasoning effort
    conditions.} Each panel shows mean reasoning token counts
    (purple bars, top axis) and accuracy (green bars, bottom
    axis) for low, medium, and high effort in \textbf{(A)}
    GPT-OSS-20B and \textbf{(B)} GPT-OSS-120B. Pearson $r$
    values (log tokens vs.\ log human RT) are annotated on each
    bar. All correlations are significant ($p < .01$) and
    remain highly stable across effort levels within each task.}
  \label{fig:barchart}
\end{figure*}

Figure~\ref{fig:barchart} displays within-task alignment
coefficients for both models across all six tasks and three
effort levels. Alignment is consistently positive and
statistically significant across every (task $\times$ effort)
combination. For GPT-OSS-20B, correlations range from
$r = .24$ to $.87$ (all $p < .01$); for GPT-OSS-120B, from
$r = .27$ to $.91$ (all $p < .01$). Within any given task,
the $r$ values remain numerically close across effort
conditions, mirroring the small variation in underlying token
counts noted in the manipulation check.

\subsection{Cross-Task Structure Is Preserved Across Effort}

\begin{figure*}[!tbp]
  \centering
  \includegraphics[width=\textwidth]{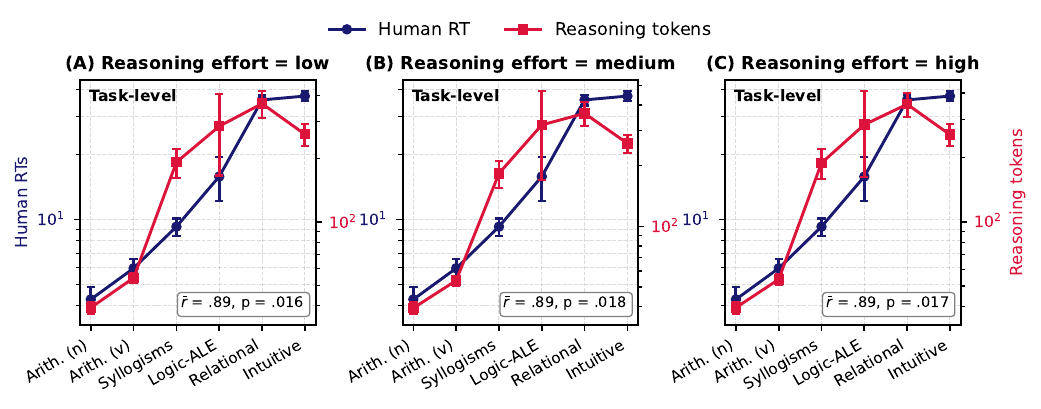}
  \caption{\textbf{Cross-task cognitive cost alignment by
    reasoning effort.} Each panel plots task-level mean human
    RTs (blue, left axis) and model reasoning tokens (red,
    right axis) on log scales. Both axes are calibrated
    independently; the close covariation of the two curves
    reflects alignment. The cross-task Pearson correlation
    $\bar{r}$ and its permutation $p$-value are inset.
    \textbf{(A)} Low, \textbf{(B)} medium, and \textbf{(C)}
    high effort yield virtually identical $\bar{r} \approx
    .89$, all $p < .02$.}
  \label{fig:cross_task}
\end{figure*}

Figure~\ref{fig:cross_task} extends the analysis to the
cross-task level. Averaging token counts across models and
items, the resulting task-level profiles mirror human RTs
with correlations of $\bar{r} = .893$ (low, $p = .016$),
$\bar{r} = .887$ (medium, $p = .018$), and $\bar{r} = .892$
(high, $p = .017$). The three effort profiles are nearly
superimposable, with the ordering of tasks from cheapest to
most expensive (Arithmetic $\to$ Syllogisms $\to$ Logic-ALE
$\to$ Relational $\to$ Intuitive) preserved identically
across conditions.

\subsection{Bayesian Tests Find No Reliable Modulation of
Alignment}

\begin{figure}[!tbp]
  \centering
  \includegraphics[width=\columnwidth]{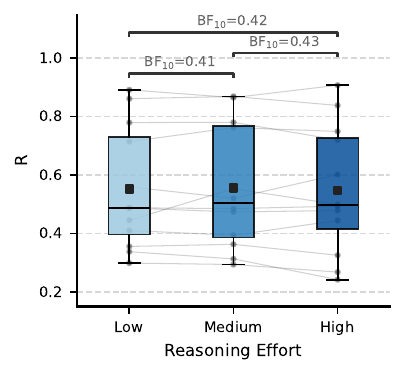}
  \caption{\textbf{Bayesian paired comparisons of within-task
    alignment ($r$) across effort levels.} Each box represents
    the distribution of Fisher $z$-transformed $r$ values
    across $n=12$ (model $\times$ task) observations. Grey
    lines connect paired observations. Bayes Factors
    ($\text{BF}_{10}$) from JZS-prior paired $t$-tests fall in
    the inconclusive range ($1/3 < \text{BF}_{10} < 1$) but
    lean toward the null in all three contrasts.}
  \label{fig:bayes}
\end{figure}

To directly test whether effort modulates within-task
alignment, we applied Bayesian paired $t$-tests to
Fisher-transformed alignment coefficients
(Figure~\ref{fig:bayes}). All three pairwise effort contrasts
yielded Bayes Factors in the inconclusive range: low vs.\
medium, $\text{BF}_{10} = 0.41$, $t(11) = -0.20$, $p = .845$;
medium vs.\ high, $\text{BF}_{10} = 0.43$, $t(11) = +0.33$,
$p = .748$; low vs.\ high, $\text{BF}_{10} = 0.42$,
$t(11) = +0.25$, $p = .811$. By the conventional thresholds
of \citet{wagenmakers2018bayesian}, these values
($1/3 < \text{BF}_{10} < 1$) do not reach moderate evidence
for the null, but they consistently lean in that direction.
More importantly, the mean alignment is numerically near
identical across conditions
($\bar{z}_\text{low} = 0.570$,
$\bar{z}_\text{med} = 0.567$,
$\bar{z}_\text{high} = 0.562$), ruling out any practically
meaningful trend even if a small statistical effect were
undetected at this sample size.

\subsection{Structural Alignment Survives Complexity Contrasts}

\begin{figure}[!tbp]
  \centering
  \includegraphics[width=\linewidth]{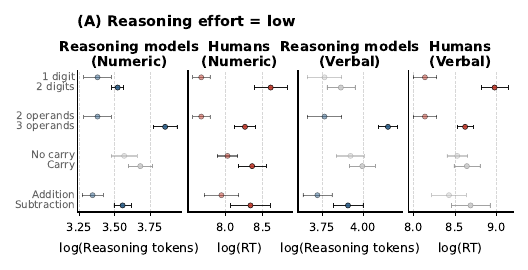}\\[4pt]
  \includegraphics[width=\linewidth]{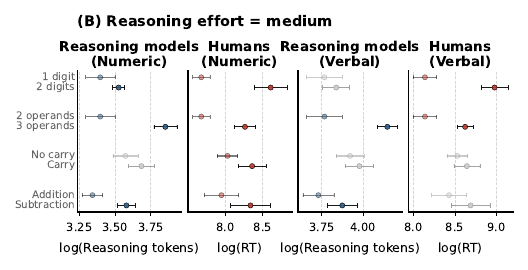}\\[4pt]
  \includegraphics[width=\linewidth]{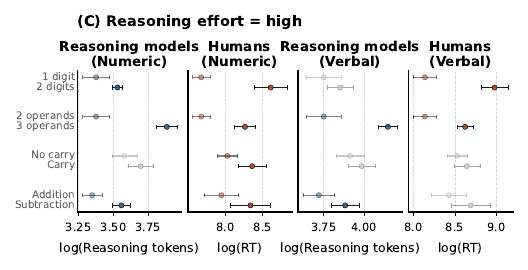}
  \caption{\textbf{Arithmetic complexity contrasts across
    reasoning effort conditions.} Each row of forest plots
    contrasts a harder vs.\ easier condition within four
    complexity dimensions. Blue points show model token
    log-means; red points show human RT log-means.
    \textbf{(A)} low, \textbf{(B)} medium, \textbf{(C)} high
    effort. Direction counts (harder $\to$ more tokens,
    $t < 0$) are reported in each panel. Human baseline: 8/8
    correct directions, 6/8 statistically significant
    ($p < .05$). The carry contrast is significant in the
    numeric format but not the verbal format in humans, an
    asymmetry that GPT-OSS-120B reproduces more faithfully
    than GPT-OSS-20B.}
  \label{fig:arithmetic}
\end{figure}

To address the concern that LRM token outputs may not track
genuine computational complexity
\citep{vankov2026correlations}, we examined whether
within-task complexity manipulations produce token profiles
directionally consistent with human RT profiles
(Figure~\ref{fig:arithmetic}). Across 16 directional
predictions per effort level (4 contrasts $\times$ 2 formats
$\times$ 2 models), models produced the correct direction of
complexity sensitivity in 16/16 cases under low effort, 16/16
under medium effort, and 15/16 under high effort. The single
exception under high effort involved the 1 vs.\ 2-digit
contrast in the verbal format for GPT-OSS-20B
($t = +0.019$, $p = .985$), a marginal reversal absent in
GPT-OSS-120B ($t = -2.07$, $p = .045$).

Beyond direction, the pattern of statistical reliability also
broadly tracks human sensitivity. Operand-count contrasts
produce robust effects in both humans and models, reaching
significance in all conditions and formats (pooled model
$p < .001$). The carry effect, by contrast, is smaller and
format-dependent in humans: significant for numeric problems
($t = -2.80$, $p = .007$) but not for verbal problems
($t = -1.13$, $p = .26$). The pooled model analysis recovers
this asymmetry directionally but at attenuated significance
(numeric $p \approx .08$ across efforts; verbal $p > .23$),
while GPT-OSS-120B alone reaches significance on the numeric
carry contrast under all three effort settings (all
$p < .03$), matching the format-dependent human pattern that
the smaller model fails to capture. This format-dependent
and scale-dependent parallel argues against a simple
verbosity-template account: token allocation tracks the
structural difficulty of the underlying computations rather
than reflecting a uniform inflation across items, and the
fidelity of this tracking improves with model scale.

\section{Discussion}

Across two model sizes, three reasoning effort levels, six
tasks, and multiple analysis approaches (manipulation checks,
within-task correlations, cross-task profiles, Bayesian
contrasts, and arithmetic complexity tests), we find no
evidence that inference-time reasoning effort modulates the
alignment between LRM token usage and human cognitive costs.
The Effort Invariance null is supported by converging
evidence: numerically near-identical mean alignment across
conditions, Bayes Factors leaning toward the null, and
preserved task-level structure. These results suggest that
human-model cognitive cost alignment is a stable structural
property of RLVR-trained reasoning rather than a
calibration-sensitive artifact.

\subsection{What the Manipulation Check Reveals}

A key finding of the manipulation check is that the
\texttt{reasoning\_effort} parameter modulates token output
only modestly, particularly for tasks with short baseline
traces. We interpret this as indicating that the parameter
sets an upper budget on generation rather than driving
real-time computational allocation. The actual allocation
across items appears determined by a policy learned during
RLVR training and is largely independent of inference-time
budget. This dovetails with the mechanistic account
proposed by \citet{hu2026h}, whose Functional Alignment
Collapse showed that SFT distillation breaks the alignment
even when surface form is preserved. Together, the two
findings sketch a picture in which human-aligned cost
allocation is a training-time achievement, robust to
inference-time perturbations of the budget but fragile to
perturbations of the training objective.

\subsection{Reconciling with Vankov et al.}

Our findings stand in partial agreement with, and partially
extend, \citet{vankov2026correlations}. Vankov et al.\
argued that the \texttt{reasoning\_effort} parameter does
not substantially affect accuracy and that models generate
verbose traces for trivial problems, concluding that token
generation reflects trained verbosity rather than dynamic
resource allocation. We confirm that effort manipulations
leave both accuracy and alignment largely unaffected, but
draw a different conclusion: the invariance reflects
insulation from inference-time noise rather than evidence
against alignment. Crucially, our complexity-contrast
analysis shows that token outputs track the direction of
human sensitivity in 47/48 cases across effort levels, and
mirror the format-dependent magnitude profile of the human
carry effect: significant for numeric problems, weak for
verbal problems, with GPT-OSS-120B reproducing this
asymmetry more faithfully than GPT-OSS-20B. This graded,
format-sensitive, scale-dependent pattern is difficult to
account for as a generic verbosity template. The example of
verbose traces for $2 + 2$ likely reflects a floor
phenomenon: models scale tokens with complexity, but the
floor is very low for trivial items \citep{de2026vankov}.

\subsection{Reconciling with Hu}

\citet{hu2026thinking} raised the concern that token-level
traces may function as performative scaffolding that
conditions subsequent generations without faithfully
representing the underlying computations
\citep{stechly2025beyond, palod2025performative}. Our
findings do not adjudicate the faithfulness question
directly, and we agree with \citet{de2026hu} that traces
need not be a transparent window onto computation any more
than verbal protocols are in humans \citep{paul2024making}.
However, our complexity-contrast results constrain how
unfaithful the traces can be. Pure scaffolding accounts
predict that token length should be either loosely coupled
to problem structure or coupled only through coarse cues
such as input length. Our results show finer-grained
coupling: the format-dependent attenuation of the carry
effect in models tracks the format-dependent attenuation in
humans, and this match improves with model scale. A purely
performative account would need to explain why scaffolding
becomes systematically more aligned with fine-grained human
cognitive structure as model capacity grows.

\subsection{Reconciling with Dujmovi{\'c}}

\citet{dujmovic2026no} argued, via a ``bird vs.\ car''
analogy, that correlated behavioral profiles between two
systems do not license inferences about shared algorithmic
mechanisms. We accept this point and follow
\citet{de2026dujmovic} in framing the alignment at the
rational rather than algorithmic level
\citep{gershman2015computational}. Our results add two
constraints to that framing. First, alignment is stable
across inference-time manipulations, which weakens the
``coincidental correlation'' reading: a coincidence between
two unrelated processes would not be expected to remain
invariant under intervention. Second, the parallel structure
extends to fine-grained within-task contrasts whose
magnitude tracks the format-dependent strength of the human
effect. This does not establish algorithmic identity, but it
does push the alignment beyond the level of coarse
behavioral covariance that motivates the bird-and-car
caution.

\subsection{Implications for Cognitive Modeling}

The picture emerging across our analyses suggests that LRMs
instantiate something closer to a compiled difficulty
representation than to an online computation. This parallels
theories of skill acquisition in which expert performance
reflects proceduralized knowledge structures relatively
insensitive to time pressure on the timescale of a single
trial \citep{anderson1982acquisition, logan1988toward}. The
model's problem-solving policy is crystallized through
thousands of RL episodes and becomes robust to inference
budget perturbations, much as a skilled human reasoner given
more time on a problem tends to produce qualitatively
similar rather than categorically different reasoning. From
this perspective, the modest effects of the
\texttt{reasoning\_effort} parameter are not a failure of
the manipulation but a property of the system: a
crystallized policy does not reorganize in response to
shallow runtime knobs \citep{mccoy2024embers}.

\subsection{Limitations and Future Work}

We tested only two models from one family (GPT-OSS), so the
invariance finding cannot yet be generalized to other
RLVR-trained LRMs such as DeepSeek-R1
\citep{guo2025deepseek}. The Bayesian tests had modest power
($n = 12$) and yielded inconclusive Bayes Factors; the
converging numerical near-identity of mean alignment
provides stronger evidence, but a fully powered replication
is valuable. Token count is also a coarse proxy for
computational cost; per-token layer depth
\citep{chen2026think} may better capture intrinsic effort.
Future work should test invariance under more extreme budget
manipulations, compare against non-LRM instruction-tuned
baselines to provide a causal control for the training
mechanism hypothesis, and probe whether the same invariance
characterizes latent-space reasoning architectures
\citep{hao2024training, goyal2024think}.





\printbibliography

\end{document}